\documentclass[5p,times,procedia]{elsarticle}
\usepackage{ecrc}
\usepackage{amsmath}
\usepackage{algorithm}
\usepackage{algorithmic}
\usepackage{booktabs}
\usepackage{tikz}
\usetikzlibrary{arrows.meta,positioning,shapes.geometric,shapes.misc,fit,backgrounds}
\tikzset{
    /.style = {
        state,
        draw=blue!60,
        fill=blue!5,
        minimum height=2.2em,
        text width=2.8cm,
        align=center,
        font=\small
    },
    /.style = {
        state,
        draw=black!60,
        fill=black!5,
        minimum height=2.2em,
        text width=1.5cm,
        align=center,
        font=\small
    },
    action_style/.style = {
        font=\scriptsize\ttfamily,
        above,
        sloped
    },
    decision_style/.style = {
        font=\small,
        align=center
    }
}

\volume{00}

\firstpage{1}

\journalname{Procedia CIRP}

\runauth{R. Fox et al.}

\jid{trpro}

\usepackage{amssymb}

\usepackage[figuresright]{rotating}

\usepackage[bookmarks=false]{hyperref}
    \hypersetup{colorlinks,
      linkcolor=blue,
      citecolor=blue,
      urlcolor=blue}

\begin{document}
\begin{frontmatter}



\dochead{11th CIRP Conference on Assembly Technologies and Systems (CATS 2026)}%

\title{State-Augmented Graphs for Circular Economy Triage}


\author[a]{Richard Fox\corref{cor1}} 
\author[a]{Rui Li}
\author[a]{Gustav Jonsson}
\author[a]{Farzaneh Goli}
\author[b]{Miying Yang}
\author[b]{Emel Aktas}
\author[a]{Yongjing Wang}

\address[a]{Department of Mechanical Engineering, University of Birmingham, Edgbaston, Birmingham, B15 2TT, UK}
\address[b]{ School of Management, Cranfield University, Cranfield, Bedfordshire, MK43 0AL, UK}

\cortext[cor1]{* Corresponding author. {\it E-mail address:} r.fox.1@bham.ac.uk}

\begin{abstract}
Circular economy (CE) triage is the assessment of products to determine which sustainable pathway they can follow once they reach the end of their usefulness as they are currently being used. Effective CE triage requires adaptive decisions that balance retained value against the costs and constraints of processing and labour. This paper presents a novel decision-making framework as a simple deterministic solver over a state-augmented Disassembly Sequencing Planning (DSP) graph. By encoding the disassembly history into the state, our framework enforces the Markov property, enabling optimal, recursive evaluation by ensuring each decision only depends on the previous state. The triage decision involves choices between continuing disassembly or committing to a CE option. The model integrates condition-aware utility based on diagnostic health scores and complex operational constraints. We demonstrate the framework's flexibility with a worked example: the hierarchical triage of electric vehicle (EV) batteries, where decisions are driven by the recursive valuation of components. The example illustrates how a unified formalism enables the accommodation of varying mechanical complexity, safety requirements, and economic drivers. This unified formalism therefore provides a tractable and generalisable foundation for optimising CE triage decisions across diverse products and operational contexts. 
\end{abstract}


\end{frontmatter}



\enlargethispage{-7mm}
\section{Introduction \& Background}
\label{intro}

Circular economy (CE) strategies aim to extend product lifecycles, reduce waste, and recover value from end-of-life (EOL) assets \cite{kirchherr2017conceptualizing}. While CE principles have gained traction across sectors \cite{ahmed2024circular}, the challenge of routing products to appropriate pathways e.g., reuse, repair, remanufacture, repurpose, or recycle, remains largely unresolved. There are more pathways reflected in the “10R” hierarchy of CE routes \cite{potting2017circular} but those listed remain the most relevant for this work. In practice, EOL routing decisions are often made heuristically or reactively, without formal models to evaluate trade-offs between retained value, disassembly cost and resource availability. This gap is particularly acute in domains where safety, regulatory compliance, and operational constraints interact, such as electric vehicle (EV) batteries \cite{baars2021circular}.

Conventional disassembly sequencing planning (DSP) methods optimise for disassembly efficiency but rarely incorporate condition-aware routing, resource constraints, or recursive evaluation of retained value versus cost \cite{wu2024knowledge}. The term triage originates from military and medical processes but is generally applicable to assessments that determine the priority of further action \cite{OED_triage}. Motivated by the context of CE triage, a framework is introduced that extends graph-based DSP by augmenting nodes with disassembly history and diagnostic information. Enabling dynamic evaluation of whether to continue disassembly or commit to a CE option. This state-augmented representation supports exact dynamic programming (DP) or reinforcement learning (RL) rollouts, bridging the gap between static optimisation and triage under real-world constraints.

EV battery disassembly at EOL is a counterexample where there are many high-value CE routes that are plausible. As such, this has been widely studied due to its critical role in safety, cost recovery, and environmental compliance. Traditional approaches to DSP have focused on optimising the order of operations under multiple objectives such as hazard reduction, energy efficiency, and economic return \cite{wu2024knowledge}. In the EV battery domain, recent work has introduced multi-objective optimisation models, probabilistic inference over disassembly graphs, and hybrid human-machine metaheuristics to improve sequence selection \cite{xiao2023multi, wu2022multi, kay2022robotic}. However, these approaches typically assume deterministic conditions and fixed routing goals, and rarely incorporate diagnostic observations or dynamic evaluation of CE options. 

While these methods provide rigorous formulations, they often lack interpretability and scalability for practitioners. Matrix-based representations, for instance, can encode precedence using constructs like stability or priority matrices \cite{xu2020disassembly, wang2021interlocking}, but these abstractions can obscure the product's physical structure. To address this, recent work has shifted toward graph-based models. Some researchers explore hybrid metaheuristics, such as genetic algorithms, to balance profit and workload \cite{cong2023retired}. Others use AND/OR graphs or dual-objective models to manage hazards and energy costs \cite{zhan2023environment,tian2018modeling}. Existing graph-based approaches often rely on heuristics to solve NP-hard formulations, our method employs a Markov decision formulation that is solvable in polynomial time and supports dynamic optimization \cite{littman2013complexity}. This structure avoids the temporal stacking or recurrence required by deep approaches for non-Markovian problems \cite{ni2022recurrent}.

This paper addresses the lack of formal triage models for CE decision-making. We aim to develop a CE triage model which considers both environmental and economic value during disassembly process. The approach formally models a continue disassembly or commit to a CE option decision, enabling planning to balance retained value against disassembly costs and operational constraints. The modelling of the economic and environment parameters, such as what determines a health score or what criteria a given CE route has in reality, is not undertaken. Instead this work intends to provide a formulation and framework for such metrics to be assessed and used effectively once modelled. We demonstrate the formulation using a representative but purely illustrative example of an EV battery module, demonstrating how the framework navigates recursive, hierarchical valuation to make triage decisions. Including realistic parameters, but with arbitrary values that serve to demonstrate how the logic works.

\section{Methodology}
We represent disassembly sequencing as a directed acyclic graph (DAG) $G=(V,E)$. Where each node $v \in V$ denotes a product, sub-assembly, or component, and each directed edge $e=(u \to v)\in E$ denotes a feasible disassembly step that produces $v$ after $u$, Figure \ref{fig:state_augmented_graph}(a). For convenience, we use the terms "node" and "state" interchangeably. A DAG allows nodes to have multiple parents, preserving shared components and enabling alternative sequences while maintaining a well-defined partial order for precedence constraints. This flexibility is essential for realistic assemblies such as battery packs or surgical instruments, where conditional steps and cross-links occur frequently.
\begin{figure}
    
\begin{tikzpicture}[
        node distance=1.5cm,
        >={Stealth[length=5pt]} 
    ]
        \node[font=\small\bfseries, above] at (0, 0.5) {(a) Standard DSP Graph ($G$)};
        
        \node[circle,fill=gray!30,behind path] (v_0) at (0,0) {$v_0$};
        \node[circle,fill=gray!30,behind path] (v_1) [below right=-0.5cm and 0.8cm of v_0] {$v_1$};
        \node[circle,fill=gray!30,behind path] (v_2) [below right=-0.5cm and 0.8cm of v_1] {$v_2$};
        \node[circle,fill=blue!30,behind path] (v_3) [below right=0.5cm and 0.8cm of v_2] {$v_3$};
        \node[circle,fill=gray!20,behind path] (v_4) [below left=1.2cm and 0.5cm  of v_2] {$v_4$};
        \node[circle,fill=orange!20,behind path] (v_5) [below left=1.2cm and -0.5cm of v_2] {$v_5$};
        
        \path[->] (v_0) edge [left] node [action_style, pos=0.8] {$e_0$} (v_1);
        \path[->] (v_1) edge [left] node [action_style, pos=0.8] {$e_1$} (v_2);
        \path[->] (v_2) edge [left] node [action_style, pos=0.8] {$e_2$} (v_3);
        \path[->] (v_2) edge [left] node [action_style, pos=0.8] {$e_4$} (v_4);
        \path[->] (v_4) edge [left] node [action_style, pos=0.8] {$e_5$} (v_5);
        \path[->] (v_1) edge [bend right=50, right] node [action_style] {$e_3$} (v_4);
        
        \node (dec2) [right=0.7cm of v_5, decision_style, text=orange] {$e_{CE}$?};
        \path[->, dashed, orange] (v_5) edge (dec2);
        \node (dec3) [right=0.5cm of v_3, decision_style, text=blue] {$e_{CE}$};
        \path[->, dashed, blue] (v_3) edge (dec3);
    \end{tikzpicture}
    

    \begin{tikzpicture}[
        node distance=1.5cm,
        >={Stealth[length=5pt]} 
    ]
        \node[font=\small\bfseries, above] at (0, 0.5) {(b) State-Augmented Graph ($\tilde{G}$)};
    
        \node[rectangle,fill=gray!30,behind path] (v0) at (-1.0,0) {$\tilde{v}_0 = (v_0, \emptyset)$};

        \node[rectangle,fill=gray!30,behind path] (v1) [below right=-0.5cm and 0.8cm of v0] {$\tilde{v}_1 = (v_1, e_0)$};
        \node[rectangle,fill=gray!30,behind path] (v2) [below right=-0.5cm and 0.8cm of v1] {$\tilde{v}_2 = (v_2, \{e_0, e_1\})$};
        
        \node[rectangle,fill=blue!20,behind path] (v3) [below left= 1.8cm and -3.0cm of v2] {$\tilde{v}_3 = (v_3, \{e_0,e_1,e_2\})$};
        \node[rectangle,fill=gray!30,behind path] (v4) [below left= 0.6cm and -0.5cm of v1] {$\tilde{v}_4 = (v_4, \{e_0,e_3\})$};
        \node[rectangle,fill=gray!30,behind path] (v5) [below left= 1.2cm and -0.5cm of v2] {$\tilde{v}_5 = (v_4, \{e_0,e_1,e_4\})$};

        \node[rectangle,fill=blue!20,behind path] (v6) [below left= 2.2cm and -0.5cm of v1] {$\tilde{v}_6 = (v_5, \{e_0,e_3,e_5\})$};
        \node[rectangle,fill=blue!20,behind path] (v7) [below left= 2.8cm and -0.5cm of v2] {$\tilde{v}_7 = (v_5, \{e_0,e_1,e_4,e_5\})$};

        \path[->] (v0) edge node [action_style, pos=0.6] {$\tilde{e}_0 \equiv e_0$} (v1);
        \path[->] (v1) edge node [action_style, pos=0.6] {$\tilde{e}_1  \equiv e_1$} (v2);
        
        \path[->] (v1) edge node [action_style, pos=0.6] {$\tilde{e}_3  \equiv e_3$} (v4);
        \path[->] (v2) edge node [action_style, pos=0.6] {$\tilde{e}_4 \equiv e_4$} (v5);
        \path[->] (v2) edge node [action_style, pos=0.6] {$\tilde{e}_2 \equiv e_2$} (v3);
        
        \path[->] (v4) edge node [action_style, pos=0.6] {$\tilde{e}_5 \equiv e_5$} (v6);
        \path[->] (v5) edge node [action_style, pos=0.6] {$\tilde{e}_6 \equiv e_5$} (v7);
        
        \node (dec) [below right=0.6 and -1.5 of v6, decision_style, text=blue] {$e_{CE}$};
        \path[->, dashed, blue] (v6) edge (dec);
        \path[->, dashed, blue] (v7) edge (dec);
        \node (dec3) [below=0.5cm of v3, decision_style, text=blue] {$e_{CE}$};
        \path[->, dashed, blue] (v3) edge (dec3);

    \end{tikzpicture}
    \caption{
        $e_{CE}$ is a possible point at which a CE option could be taken. Blue CE edges are where a decision can be made, and orange is where the node information is indistinct. In reality, all nodes are able to exit disassembly and take a CE route, but for clarity and conciseness, only terminal nodes are shown here.
        (a) A standard DSP graph $G$, 
        (b) The proposed state-augmented graph $\tilde{G}$, which 'unrolls' the paths. The states $\tilde{v}_6$ and $\tilde{v}_7$ are distinct, as their history $\tau$ differs, but they still represent the same physical node $v_5$. 
    }
    \label{fig:state_augmented_graph}
\end{figure}

The structure of $G$ alone does not guarantee the Markov property where such states must be distinct as they can only depend on the state at the previous time-step. In this work, we adopt a \emph{state-augmented DSP graph} $\tilde G = (\tilde{V},\tilde{E})$, Figure \ref{fig:state_augmented_graph}(b), using a well known technique for inducing the Markov property, state-augmentation \cite{sootla2022sauteRL, verma2023state}. The node set is augmented so that each decision state encodes not only the physical component but also the history of executed steps (edges), which will also relate to feasibility and cost. Figure \ref{fig:state_augmented_graph}(a) shows how multiple histories (e.g., $e_0 \rightarrow e_3 \rightarrow e_5$ and $e_0 \rightarrow e_1 \rightarrow e_4  \rightarrow e_5$) converge on the same physical node ($v_5$). For node $v_5$ the choice is ambiguous as the processes already performed are not captured. The proposed state-augmented graph $\tilde{G}$, 'unrolls' the paths. The states $\tilde{v}_6$ and $\tilde{v}_7$ are distinct, as their history $\tau$ differs, but they still represent the same physical node $v_5$. In turn, allowing the admissibility function to correctly approve a CE action (e.g., 'Reuse') for one state while rejecting it for another. 

By enforcing the Markov property, the framework enables exact and tractable solution methods from the simple deterministic DP solver, algorithm \ref{alg:solver}, to RL rollouts \cite{sutton2018reinforcement}. It also improves generalisability across products as the mathematical model does not depend on the particular product, instead requiring that information to be encoded in the state. Formally, let $\tilde V$ denote the lifted node set where each $\tilde{v}\in\tilde{V}$ encodes $(v,\tau)$, where $v$ is the physical component and $\tau$ is the trajectory of edges traversed on the state- augmented graph $\tau := \{\tilde e_{1}, \tilde e_{2}, \cdots, \tilde e_{t-1}\}, \: \tilde e_i \in \tilde E \; \forall i$ and represents the executed history required for option admissibility and cost accounting.  The resulting graph remains acyclic and supports standard precedence constraints, as illustrated in Figure \ref{fig:state_augmented_graph}(b), while enabling recursive scoring and dynamic optimisation. The state definition ensures that the next decision depends only on the current augmented state. This design preserves optimality under DP and enables future integration with RL, while avoiding the computational burden of global optimisation and the heuristic dependence of other non-Markovian search methods. 

We define a finite-horizon decision process on $\tilde{G}$ where leaf nodes are terminal states and CE options are always terminal states. The system occupies a state identified as a node $\tilde{v} = (v,\tau)$, and the decision-maker chooses an action $\tilde e$ from the admissible set $\mathcal{A}_{\tilde{v}} :\subseteq {\delta^{+}(\tilde{v}) \cup \mathcal{K}_{v}}$ which is the set of all outgoing edges $e \in \delta^{+}(\tilde{v})$ from node $\tilde{v}$ and all valid CE routes $k \in \mathcal{K}_v$. Therefore, two classes of actions are established: (i) primitive disassembly steps that correspond to edges $e\in\tilde{E}$; and (ii) terminal routing decisions that assign the current physical node $v$ to a CE option $k\in\mathcal{K}_v$ once access and information are sufficient. Let $\mathcal{K}$ denote the universe of CE routing options, and $k \in \mathcal{K}$ are selected based on technical diagnostics and CE feasibility. Not every option is feasible for every node; we define the node-specific admissible set of CE options $\mathcal{K}_v \subseteq \mathcal{K}, v\in V,$. Each option $k \in \mathcal{K}$ is associated with a utility $U_{v,k}$ that depends on the estimated condition of the item, the cost of processing, and the expected market value of the outcome. The triage decision for node $v$ selects exactly one option from $\mathcal{K}_v$., the one with the highest utility $\max_{k \in \mathcal{K}_v}U_{v,k}$ For node $v$ and option $k\in\mathcal{K}_v$, we define the CE options utility as:
\begin{equation}
U_{v,k}(\mathcal{H}_v,\tau) =
\mathrm{Rev}_{v,k}(\mathcal{H}_v)-\mathrm{Cost}_{v,k}(\tau),
  \label{eq:utility}
\end{equation}
where $\tau$ is the history of disassembly decisions that influence process time, consumables, and access costs. Sensors (e.g., vision, IR, vibration, diagnostics) provide health score $\mathcal{H}_v$ which could also be defined as an aggregate recursively from its components. We tie the state return $r(\tilde{v}, \tilde e)$ to the utility $U_{v,k}$. Executing a disassembly edge $e$ incurs a negative return equal to its time $t_e$- or cost $c_e$-weighted penalty. The final routing to option $k$ yields the retained value as modelled by the utility. Formally, for state $\tilde{v}$ and action  $\tilde{e}$: 
\begin{equation}
    r(\tilde{v}, \tilde e) = 
    \begin{cases} -\,c_e, & \text{if } a \text{ executes disassembly edge } e, \\  U_{v,k}\!\left(\mathcal{H}_v,\tau\right), & \text{if } a \text{ commits to CE       option } k \in \mathcal{K}_v. 
    \end{cases} 
    \label{eq:stage-reward} 
\end{equation} 
Admissible CE options depend on access and information. Let $\sigma(v)$ summarise the set of observable condition parameters available at $\tilde{v}$ and $\tau$ already encapsulates the set of executed edges required to access this state. We adopt a formulation in which the planner selects actions at each augmented state $\tilde{v} = (v, \tau)$ according to a deterministic (or stochastic) policy $\pi(\tilde{e} \mid \tilde{v}), \: \tilde e \in \mathcal{A}_{\tilde{v}}$. The expected return over the planning horizon is given by: 
\begin{equation} 
J(\tilde G) =  \sum_{\tilde{e} \in \mathcal{A}_{\tilde{v}}} \pi(\tilde{e} \mid \tilde{v}) \, r(\tilde{v}, \tilde{e}). \label{eq:on-policy-objective} 
\end{equation} 
This formulation enables recursive evaluation of disassembly paths and CE outcomes, while preserving tractability and interpretability. The policy $\pi$ may be optimised via DP and other planning algorithms, as well as more thorough and exhaustive graph-based solver methods, and the Markov property ensures that all relevant information is encoded in the current state $\tilde{v}$. While the formulation in equation \eqref{eq:on-policy-objective} expresses the expected return over all states in the augmented graph, a more precise version restricts the summation to the trajectory actually induced by the policy $\pi$. This yields a rollout-style objective: 
\begin{equation} J(\tilde G) = \mathbb{E}_{\pi} \left[ \sum_{t=0}^{T} r(\tilde{v}_t, \tilde{e}_t) \right], 
\label{eq:trajectory-objective} 
\end{equation} 
where $\tilde{v}_t$ is the state at time $t$, $\tilde{e}_t \sim \pi(\cdot \mid \tilde{v}_t)$ is the selected action, and $r(\tilde{v}_t, \tilde{e}_t)$ is the corresponding return. 

This trajectory-based formulation provides two key benefits. First, it enables dynamic rollout of the triage process, allowing the planner to evaluate disassembly paths and CE outcomes recursively. Second, it offers computational savings by avoiding evaluation over unreachable or irrelevant states, focusing only on those visited under the current policy. The formulation above provides a general framework for dynamic triage optimisation over a state-augmented DSP graph. This structure enables principled comparison between continuing disassembly and committing to a CE option, grounded in product health, uncertainty, and cost.

For the sake of brevity and clarity, we use binary decisions whereby the option with the maximal value is always chosen and allows. This choice also allows us to compute the examples sequentially by hand instead of iteratively as long as we keep the state-augmented graphs reasonably small, $\sim 10$ nodes. At any $\tilde{v}=(v,\tau)$, the decision is to \emph{stop} with some admissible option $k$ or \emph{continue} via a feasible disassembly step $e$, respecting precedence, safety, and any other constraints. 

Allowing us to define a utility function $U$: 
\begin{equation}
    U(\tilde{v})=\max\Bigg[
    \underbrace{\max_{k\in\mathcal K_v} \big[ 
    U_{v,k}(\mathcal H_v,\tau)
    \big]}_{\text{commit to CE option}},
    \underbrace{\max_{e\in \mathrm{out}(\tilde{v})} \big[ -c_e + U(\tilde{v}^{\prime})\big]}_{\text{continue disassembly}} \Bigg]. 
    \label{eq:value}
\end{equation} 
where $\tilde{v}^{\prime}$ is the state that taking edge $e$ leads to. The general approach to problem definition is now discussed which illustrates the framework's practical application and recursive structure. A simple deterministic solver (Algorithm \ref{alg:solver}) is provided, along with the admission function (Algorithm \ref{alg:admissible}) for the constraints discussed below. Then, in section \ref{sec:workedEx}, example values are assigned to perform the hierarchical triage of EVs, with emphasis on battery modules and subsystem-level recursion.

Algorithm \ref{alg:solver}'s recursive structure allows for comparison of taking the same or different routes now or later. As in, it captures the case where the resale of components is more valuable than the remanufacture or resale of the whole assembly.

\begin{algorithm}[htbp]
\caption{Triage Utility $U(v)$: Value iteration on $\tilde{G}$ (stop and take CE route vs continue disassembly)}
\label{alg:solver}
\begin{algorithmic}[1]
\REQUIRE State $v=(v, \tau)$; health map $\mathcal{H}$; option set $\mathcal{K}_v$; edge costs $c_e$
\ENSURE Optimal utility $U(v)$

\STATE $U \leftarrow -\infty$
\STATE $\mathcal{A}_{\tilde{v}}\leftarrow \phi(\tilde{v})$ \COMMENT{$\phi$ is defined in algorithm \ref{alg:admissible}}

\STATE \COMMENT{Utility of CE options}
\STATE $\mathcal{A}_{CE} \leftarrow \mathcal{K}_{v} \,\cap\, \mathcal{A}_{\tilde{v}} $
\IF{$\mathcal{A}_{CE} \neq \emptyset$}
    \STATE $U_{CE} \leftarrow \max_{k \in \mathcal{A}_{CE}} [ U_{v,k}(\mathcal{H}_{\tilde{v}}, \tau) ]$
    \ELSE 
    \STATE $U_{CE} \leftarrow -\infty$
\ENDIF

\STATE \COMMENT{Utility of continued disassembly options}
\STATE $\mathcal{A}_{e} \leftarrow \delta^+(\tilde{v}) \,\cap\, \mathcal{A}_{\tilde{v}} $
\IF{$\mathcal{A}_{e} = \emptyset$}
    \FOR{$e \in \mathcal{A}_{e}$}
        \STATE $v' \leftarrow T(v, e)$ \COMMENT{Get next state}
        \STATE $U_{e} \leftarrow U_{e} + (-c_e + U(v'))$
    \ENDFOR
\ELSE
    \STATE $U_{e} \leftarrow -\infty$ 
\ENDIF

\STATE \COMMENT{Return the utility for the optimal choice}
\STATE $U=\max(U_{CE}, U_{e})$
\STATE \textbf{return} $U$
\end{algorithmic}
\end{algorithm}

The constraints encoded by Algorithm \ref{alg:admissible} are essential to ensure feasibility and safety and can be applied as part of the function definition for $\mathcal{A}_{\tilde{v}}$, some examples include precedence, access gating, safety and regulatory, and resource constraints. As with health value and CE options the method of obtaining such rules or scoring quantities is not pursued. Instead the admissibility function provides examples of how to encode such quantities as constraints within the proposed framework. 


\begin{algorithm}[htbp] 
\caption{Admissibility Function $\phi(\tilde{v})$: Example admissible option rules on $\tilde{G}$}
\label{alg:admissible} 
\begin{algorithmic}[1] 
\REQUIRE Component $v$; history $\tau$ (set of executed edges); health map $\sigma(v) = \mathcal{H}_v$; option set $\mathcal{K}_v$; and any constraint information like required edges $e'$, paths $\tau*$, resources available $\Gamma$ and safety thresholds $S_k$. 
\ENSURE $\mathcal{A}_{\tilde{v}} \subseteq \{\mathcal{K}_v \cup \delta^+(\tilde v)\}$ (admissible options at $\tilde{v}=(v,\tau)$) 
\STATE $\mathcal{A}_{\tilde{v}} \leftarrow \varnothing$

\COMMENT{Precedence constraint}
\FOR{$e \in \delta^+(\tilde{v})$}
    \IF{$e' \in \tau$}
        \STATE $\mathcal{A}_{\tilde{v}} \leftarrow \mathcal{A}_{\tilde{v}} \cup \{e\}$
    \ENDIF
\ENDFOR

\COMMENT{Access gating constraint}
\FOR{$k \in \mathcal{K}_v$}
    \IF{$e' \in \tau$}
        \STATE $\mathcal{A}_{\tilde{v}} \leftarrow \mathcal{A}_{\tilde{v}} \cup \{k\}$
    \ENDIF
\ENDFOR

\COMMENT{Safety and regulatory constraint}
\FOR{$k \in \mathcal{A}_{\tilde{v}}$}
    \IF{$\mathrm{S}_v > \mathrm{S}_k$}
        \STATE $\mathcal{A}_{\tilde{v}} \leftarrow \mathcal{A}_{\tilde{v}} \setminus \{k\}$
    \ENDIF
\ENDFOR

\COMMENT{Resource constraint}
\STATE $R_{\tilde{v}} \leftarrow \sum_{e \in \tau}a_{e,r}$
\FOR{$e \in \mathcal{A}_{\tilde{v}}$}
    \IF{$R_{\tilde{v}} + a_{e,r} > \Gamma$}
        \STATE $\mathcal{A}_{\tilde{v}} \leftarrow \mathcal{A}_{\tilde{v}} \setminus \{e\}$
    \ENDIF
\ENDFOR

\RETURN $\mathcal{A}_{\tilde{v}}$
\end{algorithmic} 
\end{algorithm}

Applying admission rules on $\tilde{G}$ amounts to building the admission function $\mathcal{A}_{\tilde{v}}$). Algorithm \ref{alg:admissible} encodes examples for each of the rules discussed above. In this example, function access and precedence loop over all possible edges and CE routes for node $\tilde{v}$, adding valid choices to the output. Safety and resource constraints show the opposite approach of removing choices if they do not meet criteria. The implementation needed for any given process will depend on that process and its criteria. Together, these constraints ensure that the optimisation respects physical feasibility, operational limits, and safety requirements, while enabling dynamic triage decisions over the state-augmented DSP graph. 

\section{Illustrative Numerical Example}
\label{sec:workedEx}

We illustrate a minimal disassembly sequencing graph $G=(V,E)$ and its state-augmented counterpart $\tilde G$ to show how disassembly history $\tau$ gates admissible CE options and shapes utility $U_{v,k}(\mathcal{H}_v,\tau)$.
Health $\mathcal{H}_v \in [0,1]$ is modelled as a fraction of perfect health $\mathcal{H}_v = 1$, which would give the maximal return of value for a given CE route. Costs and thresholds are designed so demonstrate how the formulation is able to change its decision and critical points. Notably, cost and revenue values represent unitless utility and need not be strictly monetary. Environmental considerations can be included by incorporating carbon emissions in disassembly edges or remanufacture incentives in the net CE cost $C^{\text{proc}}$, allowing the framework to optimize for ecological outcomes. Yielding an optimisation objective of a utility for assigning $v$ to $k$ given history $\tau$:
\begin{equation}
U_{v,k}(\mathcal{H}_v,\tau)=\mathrm{Rev}_{v,k}(\mathcal{H}_v)-\Big(\sum_{e\in\tau} c_e + C^{\mathrm{proc}}_{v,k}\Big).
\end{equation}
Nodes are the Vehicle $\mathcal{V}$ battery pack $P$ and two modules $M_1, M_2$. For simplicity, and as we are not evaluating other vehicle components, we drop the vehicle nodes and just consider the battery pack and its subsequent nodes, yielding the node set $ V=\{P, M_1, M_2\}$. 

For the pack: $\mathcal K_P=\{ \mathrm{Reu}\text{(reuse)}, 
    \mathrm{Rep}\;\text{(repurpose)}, \mathrm{Rc}\;\text{(recycle)}\}$, 
with a safety gate requiring high voltage (HV) isolation: $ A_{P,k}(\tilde{v})=\mathbb{I}[e_1\in\tau] $. The selection of CE options also assumes that such a large lithium-ion battery cannot be simply disposed of and must be taken away, hence it is not a valid $k$. 
For modules $M\in{M_1,M_2}$:$\mathcal K_M=\mathrm{Reu}\text{(reuse)},
    \mathrm{Rc}\;\text{(recycle)}, \mathrm{Disp}\;\text{(Disposal)}
    \}$,
where disposal is now possible but the modules can only be reused in other battery packs and are not safe for repurposing. 

Pack recycling is admissible immediately after HV isolation, pack reuse requires isolation and a health score above $0.9$, repurpose requires the same but with the lower threshold of $0.7$ and that the thermal shield casing is removed. After isolating and extracting exactly one module, module reuse requires extraction + diagnostics + ($\mathcal{H}\ge0.8$); recycle/disposal is always admissible post-extraction.

Edges $E$ encode feasible, precedence-respecting steps (time $t_e$, cost $c_e$, resources $a_{e,r}$ with $r\in\{\mathrm{labour\text{-}min},\mathrm{fixture},\mathrm{bench}\}$): 
\begin{equation}
\begin{array}{llll}
e_{\text{iso}}{:}& \text{HV isolation} \\
& t=10,\; c=10,\; a=(10,0,0)\\
e_{\text{cov}}{:}& \text{Remove top cover (req. for everything but reusing pack)} \\ 
& t=25,\; c=25,\; a=(25,1,0)\ \ \ (\text{req.\ }e_{\text{iso}})\\
e_{\text{shield}}{:}& \text{Remove thermal shield (req. for repurposing pack)}\\
& t=12,\; c=18,\; a=(12,0,0)\ \ \ (\text{req.\ }e_{\text{cov}})\\
e_{M_1}{:}& \text{Extract }M_1 \\
& t=20,\; c=30,\; a=(20,0,0)\ \ \ (\text{req.\ }e_{\text{cov}})\\
e_{M_2}{:}& \text{Extract }M_2 \\
& t=20,\; c=30,\; a=(20,0,0)\ \ \ (\text{req.\ }e_{\text{cov}})\\
e_{M_i}^{\text{diag}}{:}& \text{Bench diagnostics for }M_i \\
& t=10,\; c=10,\; a=(0,0,10)\ \ \ (\text{req.\ }e_{M_i})\\
\end{array}
\end{equation}

We define the expected revenues and disassembly costs, Equation \ref{eq:batt_rev_costs}, enabling us to calculate Utilities for CE options. Reuse and repurpose are dependent on the node's health value, as the return from those paths is affected by the condition of the product represented by the node. Recycling has more value for modules as they are assumed to be smaller and easier to transport. Disposal's negative revenue represents a penalty to incentivise CE practice. This path is therefore only selected when other routes mount such high costs that their final utility becomes even more negative than the disposal penalty. 

\begin{equation}
\begin{aligned}
&\mathrm{Rev}_{\mathrm{Reu}}(\mathcal H)=V_{\mathrm{Reu}}\cdot \mathrm{RVR}_{\mathrm{Reu}}(\mathcal H),\quad &&C^{\mathrm{proc}}_{\mathrm{Reu}}=60,\\
&\mathrm{Rev}_{\mathrm{Rep}}(\mathcal H)=V_{\mathrm{Rep}}\cdot \mathrm{RVR}_{\mathrm{Rep}}(\mathcal H),\quad &&C^{\mathrm{proc}}_{\mathrm{Rep}}=40,\\
&\mathrm{Rev}_{Rc}^{\mathrm{P}}=50,\quad &&C^{\mathrm{proc}}_{\mathrm{Rc}}=10,\\
&\mathrm{Rev}_{Rc}^{\mathrm{M}}=60,\quad &&C^{\mathrm{proc}}_{\mathrm{Disp}}=5,\\
&\mathrm{Rev}_{Disp}=-10,\\
\end{aligned}
\label{eq:batt_rev_costs}
\end{equation}

Recycle and disposal utilities are fixed due to their value and costs being fixed and independent of product health, assuming optimal disassembly pathing. $
    U_{P,\mathrm{Rc}} = \mathrm{Rev}_{Rc}^{\mathrm{P}} -\Big(\sum_{e\in\tau} c_e +C^{\mathrm{proc}}_{\mathrm{Rc}}\big) = \boxed{30},  
     U_{M,\mathrm{Rc}} = \boxed{-15} , 
     U_{M,\mathrm{Disp}} = \boxed{-85} .$
For health-dependent routes, retained value is parametrised via residual value ratios (RVR) as per Equation \ref{eq:batt_rev_costs} here, we define RVR functions for reuse and repurpose: 
\begin{equation}
    \begin{aligned}
    \mathrm{RVR}_{\mathrm{Reu}}(\mathcal H)=\mathrm{clip}\!\left(\frac{\mathcal H-0.85}{0.15},0,1\right),\\
\mathrm{RVR}_{\mathrm{Rep}}(\mathcal H)=\mathrm{clip}\!\left(\frac{\mathcal H-0.60}{0.30},0,1\right). 
\end{aligned}
\end{equation}

Baseline values are $V_{\mathrm{Reu}}^{P}=600, V_{\mathrm{Reu}}^{M}=200, V_{\mathrm{Rep}}^{P}=250$ which would be recovered given a node in perfect health. The $RVR$ values that are functions of the health score $\mathcal{H}$ are normalised by the difference between the thresholds for admission. However the numerator will always be $>0$ as the constant value is less than the threshold. This ensures that threshold encodes an allowance to cover for the costs as well as raw returned value. We consider three diagnostic scenarios. This work stops short of computing all utilities for all nodes for all cases and instead highlights the maximal utility and any other node utilities that are noteworthy. Observations $o_{\tilde{v}}$ are assumed to consist solely of health scores $\mathcal{H}_{\tilde{v}}$ and are: 

$\textbf{Case A (high health)}: \mathcal H_{P}(o_{P})=0.92$,

$\textbf{Case B (moderate)}: \mathcal H_{P}(o_{P})=0.82, \mathcal H_{M_1}(o_{M_1})=0.92, \mathcal H_{M_2}(o_{M_2})=0.72.$,

$\textbf{Case C (degraded)}: \mathcal H_{M_1}(o_{M_1})=0.30, \mathcal H_{M_1}(o_{M_1})=0.16, H_{M_2}(o_{M_2})=0.54.$
\subsubsection{Case A}
We can intuitively see that reuse will be the best option for this case, as it is valid and any further disassembly would only add costs. 
\begin{equation}
    \begin{aligned} 
    \mathrm{Rev}_{\mathrm{Reu}}&=V_{Reu}\;\mathrm{RVR}_{\mathrm{Reu}}(0.92)  \\
    &=600 \times \tfrac{0.92-0.85}{0.15}=280.\\
    U_{P,\mathrm{Reu}}&= \mathrm{Rev}_{Reu}^{\mathrm{P}} -\big(c_{e_{\mathrm{iso}}} +C^{\mathrm{proc}}_{\mathrm{Reu}}\big)\\
    &=280-(10+60)=\boxed{210},
    \end{aligned} 
\label{eq:caseA_util}
\end{equation}
This remains true even for repurpose with a lower CE route cost and the maximal value for the repurpose route, $\mathrm{Rev}_{\mathrm{Rep}}=V_{Rep}\;\mathrm{RVR}_{\mathrm{Rep}}(0.92)
=250 \times 1=250$, being met. It would accumulate even more disassembly costs to get there, however, $U_{P,\mathrm{Reu}}=250-\big(c_{e_{\mathrm{iso}}} +c_{e_{\mathrm{cov}}} +c_{e_{\mathrm{shield}}} +40\big) = 250 - 103 = \boxed{147}$. The logic of which holds for further disassembly's too with module extraction and CE routing incurring even more costs. Some costs would now be duplicated for each module as well as having a lower baseline value.

\subsubsection{Case B}
In this case we still have relatively high health value for the pack, but we also have health values for the constituent modules. In fact, the pack health is set to be the average of module health as an example that health scores can be aggregations of the health scores of their constituent components. Although, a simple average is too naive a metric to provide an accurate aggregate in real-world scenarios.

It is not apparent intuitively which CE options are likely to return the most value here, so we drop the listing of edges in the trajectory in the calculations as give the utilities for all variable utility nodes below:
\begin{equation}
\begin{aligned}
    U_{P,\mathrm{Reu}}\:& \text{is not valid as} \: \mathcal{H}_{P}<0.9 \therefore \mathrm{Reu} \notin \mathcal{A}_{\tilde{v}=P}\\
    U_{P,\mathrm{Rep}}&=183-113=\boxed{70}\\
    U_{M_1,\mathrm{Reu}}&=250-(10+25+30+10+60)=\boxed{115}\\
    U_{M_2,\mathrm{Reu}}\: &\text{is not valid as} \: \mathcal{H}_{M_2}<0.8 \therefore \mathrm{Reu} \notin \mathcal{A}_{\tilde{v}=M_2}\\
\end{aligned}
\label{eq:caseB_util}
\end{equation}
Reuse of module $M_1$ has by far the most utility but it must be taken in consideration with all the other module utilities to be accurately assessed at the pack level. For the return of algorithm \ref{alg:solver} we need to compute $U(\tilde{v} = P)$ as the best option for module $M_2$ is to recycle,
\begin{equation}
    \begin{aligned}
        U(\tilde{v} = P) &= \max{\big(U_{P,Rep},(U_{M_1,Reu}+U_{M_2,Rc})\big)} \\
        &= \max{\big(70, 115 + (-25)\big)}\\
        &= \max{\big(70, 90)\big)} = 90.
    \end{aligned}
\end{equation}
Recycling the other module has a utility of $-25$ and if we consider both of the modules together we receive a total return of $90$ which is still better than repurposing the whole pack, but not much and the addition of more costs or a lower health score for module $M_1$ could quickly have repurposing the whole pack becoming the optimal route.

\subsubsection{Case C}
No health-dependent routes are able to be taken in this case. Meaning we simply want to recycle the whole pack as that has the best utility of $30$ from our fixed utilities. This result highlights a key insight of the framework: while this is the economically optimal decision for the operator, it is environmentally suboptimal as it fails to separate materials. This demonstrates a clear misalignment where the fixed utility for module-level recycling $U_{M,Rc}=-15$ is too low. As will be discussed further, this suggests a clear, quantifiable target for policy interventions, such as subsidies, to make deeper, more valuable disassembly profitable.
\section{Discussion}\label{sec:discussion}
The worked examples demonstrate the framework's flexibility, but also highlight areas for refinement and discussion. In our current formulation, the health of an assembly, $\mathcal{H}_v$, is aggregated from its components using a monotone, permutation-invariant function, such as a weighted sum or minimum. While this provides a clear and computable metric, this approach may be overly strict in some contexts. For instance, a 'minimum' aggregator would condemn an entire assembly for a single faulty-but-replaceable component. Future work could explore more flexible aggregation functions that better capture the nuances of repairability. The EV battery case demonstrated the recursive, hierarchical nature of the 'Stop vs. Continue' decision, evaluating utility at the pack level versus the module level. Guiding the solver to prune inadmissible paths and select the optimal feasible CE route. 

Our adoption of a state-augmented graph enforces the Markov property, enabling DP and recursive rollouts suitable for triage. This is crucial for decisions that depend on unfolding diagnostic information. However, for processes that are highly static, well-modelled, and involve relatively few steps, this approach may introduce unnecessary overhead. In such cases, a global optimisation formulation, such as a Mixed-Integer Linear Program (MILP) \cite{klein2024mixed}, could be a more direct alternative, though it sacrifices step-by-step adaptability.

The state augmentation duplicates physical nodes across distinct access and information histories to enforce the Markov property. This increases the number of decision states, but it yields three benefits that are essential in our setting. First, it guarantees the principle of optimality, enabling exact DP, RL or other solvers without ad hoc lookahead heuristics. It also preserves a clean separation between valuation and sequencing: the same reward construction equation \eqref{eq:stage-reward} applies regardless of how the augmented graph is generated. In practice, growth in the augmented state space can be controlled by merging states with equivalent trajectory signatures and pruning dominated states based on resource or precedence bounds. 

The framework can also highlight potential economic misalignments that hinder CE goals. In our EV battery example, the optimal decision in Case C (degraded health) was to recycle the entire pack ($U_{P,Rc} = 30$) , as the utility of disassembling and recycling individual modules was strongly negative ($U_{M,Rc} = -25$). This outcome is economically rational for the operator but environmentally suboptimal, as it fails to separate materials at the module level. By formally modelling this utility gap, our framework identifies a clear intervention point where subsidies could be targeted at the module recycling step.

\section{Conclusion \& Future Work}\label{sec:conclusion}
Existing DSP literature primarily addresses the structural and operational aspects of disassembly \cite{wu2024knowledge, zhan2023environment}, with limited attention to the informational and economic dimensions critical for CE decision-making. Our contribution builds on this trajectory by introducing a state-augmented DSP graph $\tilde{G}$, where each node encodes both the physical component and its disassembly history. This design enforces the Markov property, allowing DP and RL-compatible rollout while preserving precedence and resource constraints. This approach generalises beyond EV batteries and provides a unified formalism for CE triage that is both interpretable and computationally tractable.

Future work will proceed in several directions. Extending the model to explicitly incorporate uncertainty, moving beyond the current deterministic model to include stochastic process outcomes and the "fuzziness" inherent in diagnostic data \cite{ye2022self}. 
Modelling uncertainty makes the problem an ideal fit for sequential decision-making AI. RL is designed to optimise policies in exactly these uncertain environments \cite{sutton2018reinforcement} and is a more direct fit for the decision-making task than large-scale visual or language models. However, those models show great promise for the task of condition assessment (i.e., generating the health score $\mathcal{H}$) and represent a separate avenue for future work \cite{simaei2024ai}. 

Advanced techniques, such as RL with Human Feedback (RLHF), could be used to align the learned utility functions with the nuanced preferences of expert human operators \cite{verma2023state,CalvoFullana2024state}. A key avenue for this work is to incorporate humans directly into the process, allowing assessment criteria to be learned from and controlled by experts. Another priority will be to incorporate more realistic data from industrial partners to model real-world scenarios. This will enable us to refine the framework, demonstrating its value in conjunction with businesses and highlighting its applicability across various industrial sectors.

The long-term vision is to extend the framework to multi-agent coordination. This would enable the system to optimise triage decisions across multiple workstations or facilities, managing shared resource constraints in a mixed-automation environment that includes both human and robotic agents.

\bibliography{references}
\bibliographystyle{elsarticle-num} 

\end{document}